\documentclass{article} 
\usepackage[final]{colm2026_conference}

\usepackage{microtype}
\usepackage{hyperref}
\usepackage{url}
\usepackage{booktabs}

\usepackage{lineno}

\usepackage{colortbl}
\usepackage{enumerate}
\usepackage{adjustbox}
\usepackage{enumitem}
\usepackage{booktabs}
\usepackage{xspace}
\usepackage{fontawesome5}
\usepackage[table]{xcolor}
\usepackage{collcell} 

\usepackage{amsmath,amsfonts,bm}

\def\eqref#1{equation~\ref{#1}}

\def\1{\bm{1}}

\DeclareMathAlphabet{\mathsfit}{\encodingdefault}{\sfdefault}{m}{sl}
\SetMathAlphabet{\mathsfit}{bold}{\encodingdefault}{\sfdefault}{bx}{n}

\usepackage{listings}
\usepackage{hyperref}
\usepackage{url}
\usepackage{amssymb}
\usepackage{pifont}
\usepackage{bbding}
\usepackage[most]{tcolorbox}
\usepackage{soul}

\usepackage{wrapfig}

\usepackage{tabularx}
\usepackage{array}
\usepackage[table]{xcolor}
\usepackage{booktabs}
\usepackage{pifont}
\usepackage{makecell}

\usepackage[T1]{fontenc}
\usepackage[utf8]{inputenc}

\newcommand{\headernodot}[1]{\smallskip\noindent\textbf{#1}}
\newcommand{\header}[1]{\headernodot{#1.}}

\definecolor{Gainsboro}{rgb}{0.86, 0.86, 0.86}

\definecolor{backred}{RGB}{255, 190, 190}
\definecolor{backblue}{RGB}{220, 230, 250}
\definecolor{backgreen}{RGB}{220, 250, 230}
\definecolor{backpurple}{RGB}{178, 147, 201}
\newtcbox{\bluetab}{on line, rounded corners, box align=base, colback=backblue, colframe=white, size=fbox, arc=3pt, before upper=\strut, top=-2pt, bottom=-4pt, left=-2pt, right=-2pt, boxrule=0pt}
\newtcbox{\redtab}{on line, box align=base, colback=backred, colframe=white, size=fbox, arc=3pt, before upper=\strut, top=-2pt, bottom=-4pt, left=-2pt, right=-2pt, boxrule=0pt}
\newtcbox{\greentab}{on line, box align=base, colback=backgreen, colframe=white, size=fbox, arc=3pt, before upper=\strut, top=-2pt, bottom=-4pt, left=-2pt, right=-2pt, boxrule=0pt}
\newtcbox{\purpletab}{on line, box align=base, colback=backpurple, colframe=white, size=fbox, arc=3pt, before upper=\strut, top=-2pt, bottom=-4pt, left=-2pt, right=-2pt, boxrule=0pt}

\newtcolorbox{EnvBox}[1][]{
  colback=white,
  colframe=blue!70!black,
  coltitle=white,
  fonttitle=\bfseries,
  title=#1,
  enhanced,
  breakable,
  sharp corners,
  width=\linewidth      
}

\newcommand{\ours}{\textsc{OpenReward}\xspace}

\sethlcolor{red!10}

\definecolor{darkblue}{rgb}{0, 0, 0.5}
\hypersetup{colorlinks=true, citecolor=darkblue, linkcolor=darkblue, urlcolor=darkblue}

\title{\ours: Enabling Agentic Reward Models for Long-Form Knowledge-Intensive Tasks}

\author{Ziyou Hu$^{1*}$ \quad  Zhengliang Shi$^{2}$\thanks{Equal contributions} \quad Minghang Zhu$^{2}$ \quad Haitao Li$^{3}$ \quad 
\quad Teng Sun$^2$  \\
\textbf{Pengjie Ren$^2$} \quad \textbf{Suzan Verberne$^{1}$} \quad \textbf{Zhaochun Ren}$^{1}$\thanks{Corresponding author}\\
$^{1}$Leiden University, Leiden, Netherland \quad
$^{2}$Shandong University, Qingdao, China\\
$^{3}$Tsinghua University, Beijing, China\\
\texttt{\{retrro.hu, zhengliang.shii\}@gmail.com} \\
\texttt{\{s.verberne,z.ren\}@liacs.leidenuniv.nl}
}

\begin{document}

\ifcolmsubmission
\linenumbers
\fi

\maketitle

\begin{abstract}
Reward models (RMs) are essential for aligning large language models as scalable proxies for human evaluation in both training and inference. However, existing RMs struggle in open-ended tasks with knowledge-intensive and long-form responses, where assessing correctness often requires grounding beyond the model's internal knowledge.
To tackle this challenge, we propose \ours, a type of tool-augmented RM that adaptively reasons and calls external tools to gather relevant evidence, making reliable judgments.
We train \ours using Group Relative Policy Optimization (GRPO) on over 27K pairwise data collected via a novel controllable data synthesis framework. 
The training objective jointly supervises intermediate tool usage and final judgment accuracy, incentivizing the reward model to learn evidence-based judgment strategies.  
Extensive experiments across three newly collected datasets and two widely used benchmarks demonstrate that \ours substantially outperforms existing RMs.
Furthermore, we find that integrating \ours into both inference-time response selection and preference alignment yields consistent improvements across a range of downstream tasks.
This highlight the potential of tool-augmented reward models for reliable evaluation and post-training alignment.
Our code is available on \href{https://anonymous.4open.science/r/OpenReward-F188/}{\faGithub\textbf{ }Anonymous Github}.
\end{abstract}

\section{Introduction}
\label{sec:introduction}

Reward models (RMs) have emerged as scalable and effective substitutes for human evaluators, playing a key role in aligning large language models (LLMs) during both inference and training~\citep{guo2025reward,chen2025rm}.
By learning to predict human preferences over model outputs, RMs facilitate response re-ranking at inference time~\citep{setlur2024rewarding} and provide reliable supervision signals to guide reinforcement learning (RL)~\cite{song2024preference,ouyang2022training}. 
Existing RMs are trained either as binary preference classifiers~\cite{liu2024skywork} or using the LLM-as-a-Judge~\cite{li2025generation,bercovich2025llama} paradigm.

Despite their empirical success, these approaches still face significant challenges when evaluating knowledge-intensive long-form outputs, such as the content generated by Deep Researcher systems~\citep{zheng2025deepresearcher,zhu2025deepreview}.
Evaluating such content often requires aggregating and verifying information from diverse external sources, which exceeds the internal knowledge capacity of standard reward models.
For example, evaluating the correctness of a medical report often depends on cross-referencing external scientific corpora (\textit{e.g.,} PubMed)~\citep{zhu2025deepreview}, while comparing travel itineraries demands up-to-date information of specific destinations~\citep{chen2024travelagent}. 
Most existing work either prompts commercial LLMs (\textit{e.g.,} GPT-4) to perform fact-checking via search engines~\citep{wei2024long}, or uses simple tool interactions for short-form verification (\textit{e.g.,} checking date via a calendar)~\citep{li2023tool}. 
However, these work exhibits limited flexibility and scalability, and their effectiveness in long-form reward modeling remains largely unexplored.
This gap leads to the following research question: 
\textbf{How can we design reward models that effectively evaluate long-form outputs in open-domain settings?}

In this paper, we aim at \textbf{empowering RMs to use external tools on demand in evaluation}.
Similar to task-solving agents~\citep{jin2025search,wang2024executable}, tool-augmented RMs can retrieve and reason over external information sources in real time, enabling them to make more accurate, context-aware judgments.
Building on this intuition, we propose \ours, a reward modeling framework to assess complex, knowledge-intensive, long-form responses with the assistance of external tools. 
Given two candidate responses, \ours first plans and executes a sequence of tool calls to retrieve supporting evidence, then verifies each response against this evidence, and finally selects the better answer.
To train such open-source RMs, we use Group Relative Policy Optimization (GRPO)~\citep{shao2024deepseekmath}, a widely used reinforcement learning (RL) algorithm.
The training objective combines two complementary signals: (i) an intermediate tool‑use penalty that discourages irrelevant or incorrect calls while encouraging task‑appropriate tool selection; and (ii) a final outcome‑accuracy reward that incentivizes reliably comparing the two responses and choosing the higher‑quality one.

A key obstacle in training long-form RMs is the lack of reliable pairwise response data from real-world sources, as demonstrated in Table~\ref{tab:statistics}. 
To address this, we propose a simple yet controllable partial synthesis framework named \textsc{OpenRM-Lib}, which  automatically generates queries and constructs positive–negative pairs. 
Specifically, we first perform target-aware query generation, prompting a strong LLM (\textit{e.g.,} DeepSeek-V3) to formulate a self-contained query from a sampled document. 
Then, we prompt an LLM to answer the query with and without access to the reference document, yielding a positive and a negative response, respectively. This method naturally ensures scalable synthesis of the partial order of two responses.

As a further step, we explore the utility of \ours in scaling LLM alignment at both training and inference stages. 
At inference time, we show that \ours can select a better response among candidates, improving long-form answer quality.
At training time, we employ \ours as a data selector, comparing task-solving trajectories sampled from LLMs and filtering them with \ours, thereby obtaining a higher-quality training dataset.
Our experiments demonstrate that LLMs trained with \ours-selected data achieve significant performance gains compared with those trained on data filtered by existing RMs, highlighting the potential of tool-augmented reward modeling.

Our contributions are as follows:
(i) We introduce \ours, a type of RM that can judge long-form responses with the assistance of external tools, providing accurate reward for complex agentic tasks.
(ii) We propose a simple but controllable strategy for reward data construction, and collect 27K+ high-quality training data with binary labels in an unsupervised manner.
(iii) Experiments on five benchmarks show that \ours outperforms existing baselines substantially.
(v) We demonstrate that \ours improves LLM  on both inference time and the alignment stage, facilitating future research on LLM post-training.
\section{Related Work}\label{sec:related-work}
\subsection{Reward Modeling}
Reward modeling serves as a crucial bridge connecting human intent with model behavior~\cite{li2024llmsasjudgescomprehensivesurveyllmbased,wang2024secrets}, providing supervision or judgment on model outputs.
Existing research mainly explores two paradigms of reward modeling: (i) Scalar Reward Models (Scalar RMs);
and  (ii) Generative Reward Models (\textit{a.k.a.,} LLM-as-a-Judge).
The Scalar RMs typically involve training a sequence classifier on top of a frozen LLM, training the model to predict banal human preferences or rating labels~\citep{wang2024helpsteer2,liu2024skywork}.
The later generative RMs retain the generative nature of LLMs, leveraging LLMs' language understanding and reasoning abilities to produce preference judgments~\citep{li2024llms,zheng2023judging,gu2024survey}.
Compared to traditional scalar RMs, generative RMs offer fine-grained natural language explanations, coming up with significant advantages in expressiveness, flexibility, and reliability~\cite{li2025generation}. 
Therefore, in this work, we focus on addressing the limitation of the generative reward model, \textit{i.e.,} its lack of knowledge for judging long-form, complex outputs.

\subsection{RL with Verifiable Reward}

Reinforcement learning with verifiable reward (RLVR) has recently emerged as a foundational paradigm for improving the reasoning capabilities and robustness of LLMs~\cite{qian2025toolrl,zhang2025rlvmr,zeng2025simplerl}.
By leveraging automatically verifiable criteria as reward signals, RLVR reduces dependence on subjective human preferences and helps mitigate issues such as reward hacking~\cite{xie2025capo,wen2025reinforcement,liu2025trust}.
A growing body of research has applied RLVR to the training of LLM-as-a-Judge systems, aiming to encourage deeper reasoning and more accurate evaluations through outcome-driven rewards~\cite{chen2025judgelrm,whitehouse2025j1incentivizingthinkingllmasajudge}.
For example, JudgeLRM~\cite{chen2025judgelrm} combines format and correctness rewards, demonstrating promising performance.
Similarly, J1~\cite{whitehouse2025j1incentivizingthinkingllmasajudge} transforms both verifiable and non-verifiable prompts into judgment tasks with verifiable reward signals, using RLVR to foster thoughtful reasoning while reducing systematic biases.
In contrast to prior work, our work trains a tool-augmented reward model via RLVR on automatically synthesized data.

\begin{figure}[t]
    \centering
    \includegraphics[width=\linewidth]{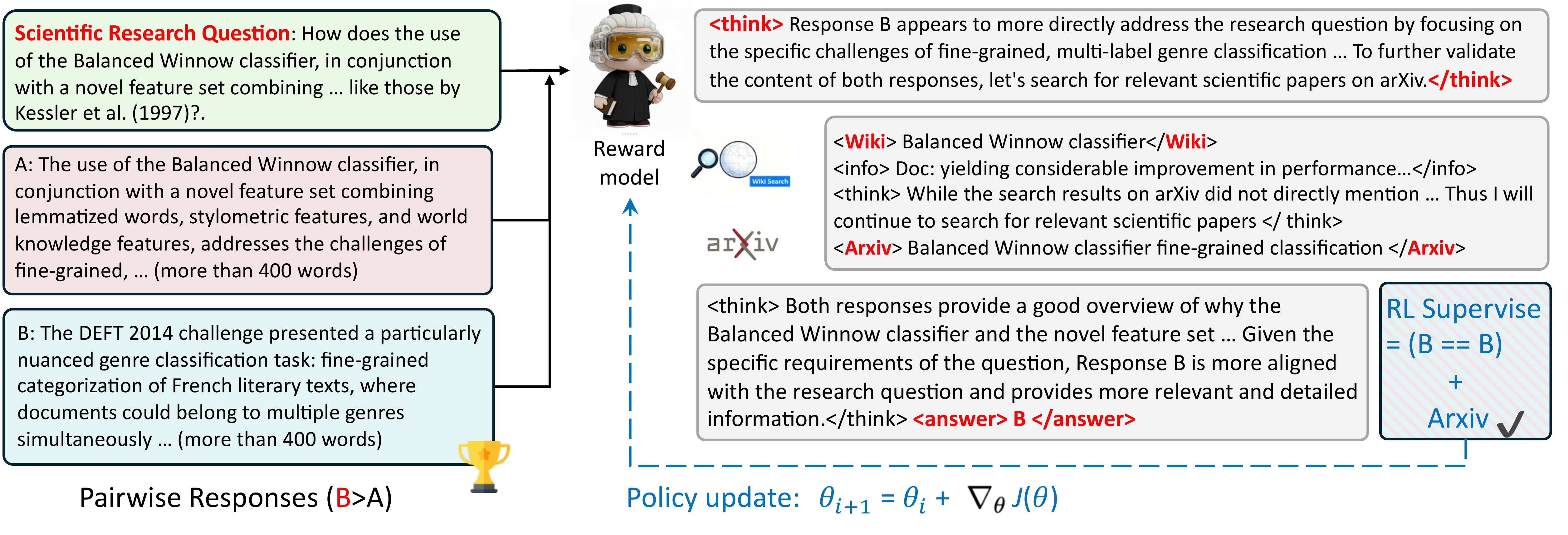}
    \caption{Illustration of the \ours framework where the reward model, when receiving the candidate responses, progressively invokes external tools to gather useful evidence, and then makes the final judgment.}
    \label{fig:flow}
    \vspace{-0.3cm}
\end{figure}

\section{Method}\label{sec:method}

This section details the proposed \ours.
We begin by explaining how \ours evaluates two candidate responses with the assistance of external tools.
Then, we introduce how such agentic judgment ability can be incentivized via RLVR.
Finally, we describe our controllable data synthesis strategy for generating large-scale training pairs.

\subsection{Rewarding Agentic Tasks with Tools}
\ours is a tool-augmented reward framework for knowledge-intensive long-form responses, where the answer quality depends on information beyond the model's internal parameters.
As illustrated in Figure~\ref{fig:flow}, \ours flexibly invokes external tools to seek necessary information, enabling accurate and evidence-grounded judgments.
Specifically, given an input query $q$ and two candidate responses $x_1$ and $x_2$, the model iteratively decides which tool to invoke to verify the information contained in the responses. The tool selection at step $i$ can be formulated as:
\begin{equation}
t_i = \mathcal{RM}(t_i \mid P, (x_1, x_2, q); \theta),
\end{equation}
Here $P$ denotes the RM' system prompt, specifying the available toolset (See Appendix~\ref{sec:app:details} for details).
$t_i$ indicates the tool selected at step $i$ and $\theta$ is the RM's parameters.
The selected tool $t_i$ is executed to obtain an external result $e_i$, which is appended to the RM's context to guide subsequent reasoning and tool selection.
This process can be formalized as a sequential decision process: the state encodes the current context, including the query, candidate responses, and prior evidence; the action corresponds to selecting a tool or terminating with a final decision; and the transition updates the context with the result $e_i$ returned by the selected tool.
The overall judgment trajectory $o$ can be summarized as:
\begin{equation*}
\small
   (q, x_1, x_2) \rightsquigarrow t_1 \xrightarrow{\text{Exec}} e_1 \dots  \rightsquigarrow t_i \xrightarrow{\text{Exec}} e_i \rightsquigarrow \dots  y.
\end{equation*}
Here $y$ denotes the final judgment, which selects the better response between $x_1$ and $x_2$ (\textit{e.g.,} \texttt{<answer> A </answer>}), accompanied by a brief explanation for interpretability purposes.
We set the maximal tool calling to $n$ during the practice, and thus the $o = \{(t_i, e_i) \mid i\in [n]\} \cup \{y\}$.

\subsection{Incentivizing Tool-augmented Judgment}\label{sec:training}

Enabling the agentic rewarding process in \ours is challenging. A straightforward approach is to directly prompt the RM to select tools and make judgments. However, this strategy may result in limited performance, particularly for smaller models such as Qwen-7B, which tend to struggle with multi-step reasoning in the absence of explicit supervision.
Another alternative is to train the model using Supervised Fine-tuning (SFT) or Rejection Sampling Fine-tuning ~\cite{li2023generative,kim2023prometheus}.
However, it passively enforces a model to imitate fixed trajectories, restricting generalization beyond handcrafted templates. 

To overcome these challenges, we adopt a reinforcement learning framework that enables LLMs to explore tool-use strategies actively. It requires only a final preference label for supervision, while encouraging the discovery of practical multi-step reasoning through trial-and-error.

\header{Training Signal}
We design a composite reward function that supervises both the intermediate tool-use behavior and the final prediction outcome.
Specifically, the reward consists of two components.
First, $\mathcal{R}_{\text{tool}}$ evaluates the accuracy of tool selection before the final decision. 
It incentivizes the model to use appropriate tools for the task and penalizes unnecessary or irrelevant tool use. 
Second, $\mathcal{R}_{\text{EM}}$ measures the correctness of the final prediction by checking whether the model's output $a_i$ exactly matches the ground-truth label, i.e., $\text{EM}(a_i)$.
The final reward is defined as:
\begin{equation}
    \mathcal{R} = \mathcal{R}_{\text{EM}} +  \text{sign}(\mathcal{R}_{\text{EM}}) \cdot \lambda \cdot \mathcal{R}_{\text{tool}},
\end{equation}\label{eq:reward}
where $\text{sign}(u)$ is set to 1 if and only if $u > 0$ and $\lambda$ is a weighting factor that balances the two reward components.
$\mathcal{R}_{\text{EM}}$ enforces the correctness of the final outcome, while $\mathcal{R}_{\text{tool}}$ provides intermediate feedback during the reasoning process by assigning $+1$ to correct tool selections and $0$ to incorrect ones.
To ensure correct judgment remains the primary learning objective, the reward function $\mathcal{R}$ only assigns credit for tool usage when the final prediction is accurate.
This design strikes a balance between promoting accurate reasoning and encouraging effective tool-use behavior.

\header{Training Process}
We train the RM with Group Relative Policy Optimization (GRPO)~\citep{shao2024deepseekmath}. 
Formally, for each query $(q, x_1, x_2)$, \ours generates a group of $m$ candidate responses using different sampled trajectories of tool use. Each trajectory $\mathcal{T}_i$ is evaluated by our composite reward function $\mathcal{R}$, which jointly considers both the intermediate tool usage quality and the final answer correctness. GRPO then computes a group-relative advantage by comparing each trajectory's reward to the mean within its group:
\begin{equation}
    \mathcal{A}_i =
    \frac{\mathcal{R}(\mathcal{T}_i) - \frac{1}{m} \sum_{j=1}^m \mathcal{R}(\mathcal{T}_j)}
         {\text{std}(\{\mathcal{R}(\mathcal{T}_j) \mid j \in [m]\})},
\end{equation}
where $\text{std}(\cdot)$ is the standard deviation of group rewards.
The final objective $\mathcal{J}(\theta) $ is a clipped policy gradient with KL regularization:
\begin{equation*}
\begin{aligned}
   \mathbb{E}_{o_i\sim \mathcal{O}} \left[
        \min\!\left(
            \rho_{i} \,\mathcal{A}_i,\;
            \text{clip}\!\left(\rho_i, 1 \pm \epsilon\right) \mathcal{A}_i \right)
          - \beta \text{KL}(\theta || \theta_{\text{ref}}) \right].
\end{aligned}
\end{equation*}
Here $\rho_{i} = \frac{\pi_{\theta}(o\mid x, \mathcal{P})}{\pi_{\text{old}}(o\mid x, \mathcal{P})}$ is the important ratio between the updated and old model parameter, $\epsilon$ controls the clipping range and $\beta$ weights the penalty for diverging from a reference model $\theta_{\text{ref}}$.

\begin{table}[!t]
\centering
\scriptsize
\setlength{\tabcolsep}{3pt}
\renewcommand{\arraystretch}{0.95}

\newcolumntype{Y}{>{\centering\arraybackslash}X}

\begin{tabularx}{0.98\columnwidth}{@{}p{2.8cm}YYYY@{}}
\toprule
\textbf{Statistic}
& \makecell[c]{\textbf{\textsc{OpenRM-Lib}} \\ (Ours)}
& \makecell[c]{\textbf{RewardBench} \\ (\citeauthor{lambert2024rewardbench})}
& \makecell[c]{\textbf{TARA} \\ (\citeauthor{li2023tool})}
& \makecell[c]{\textbf{JudgeLRM} \\ (\citeauthor{chen2025judgelrm})}
\\
\midrule

\rowcolor{green!12}
\# Training data
& \ding{52}
& \ding{56}
& \ding{52}
& \ding{52}
\\

\quad - \textit{Data Scale}
& 27K
& -
& 13K
& 100K
\\

\quad - \textit{Response Tokens}
& 582.45
& -
& 49.04
& 117.60
\\

\midrule

\rowcolor{blue!12}
\# Evaluation data
& \ding{52}
& \ding{52}
& \ding{52}
& \ding{52}
\\

\quad - \textit{Data Scale}
& 2K
& 2K
& 1.5K
& 5K
\\

\quad - \textit{Response Tokens}
& 601.03
& 93.31
& 52.19
& 116.09
\\
\bottomrule
\end{tabularx}

\caption{Basic statistics of our benchmark.}
\label{tab:statistics}
\vspace{-0.3cm}
\end{table}

\begin{table*}[!t]

\vspace{0.1in}
\vspace{0.1in}

\centering
\setlength\tabcolsep{5.0pt}
\begin{adjustbox}{width=0.97\textwidth,center}
\begin{tabular}{@{}p{9cm}|cccc@{}}
\toprule
\textbf{Domain}
& \textbf{Wikipedia ($\uparrow$)} 
& \textbf{Scientific ($\uparrow$)} 
& \textbf{Medical ($\uparrow$)} 
& \textbf{Average ($\uparrow$)}
\\
\midrule
\multicolumn{5}{@{}l}{\textit{LLM-as-a-Judge.}} \\
\midrule
Deepseek-V3.1~\citep{liu2024deepseek}
& 75.00
& 46.00
& 33.00
& 51.33
\\
GPT-4o~\citep{hurst2024gpt} 
& 70.00
& 48.20
& 44.00
& 54.07
\\
Gemini-2.5-Pro~\citep{comanici2025gemini} 
& 72.20
& 46.60
& 36.00
& 51.60
\\
Claude-Opus-4-1-20250805~\citep{anthropic_claude_2024} 
& 74.60
& 49.20
& 51.10
& 58.30
\\
Skywork-Reward-Gemma-2-27B~\citep{liu2024skywork} 
& 45.20
& 55.40
& 47.74
& 49.45
\\
JudgeLRM-7B~\citep{chen2025judgelrm} 
& 50.80
& 50.60
& 48.44
& 49.94
\\
RRM-7B~\citep{guo2025reward} 
& 56.90
& 52.95
& 53.10
& 54.32
\\
RM-R1-Qwen2.5-Instruct-7B~\citep{chen2025rm} 
& 55.40
& 54.80
& 52.30
& 54.17
\\
RM-R1 \textit{w/ continue training on our} \textsc{OpenRM-Lib-27K}
& 66.00
& 73.00
& 65.00
& 68.00
\\
\midrule
\multicolumn{5}{@{}l}{\textit{Agentic Reward Modeling.}} \\
\midrule
Deepseek-V3.1~\citep{liu2024deepseek}
& 77.00
& 48.00
& 34.00
& 53.00
\\
GPT-4o~\citep{hurst2024gpt} 
& 76.40
& 58.60
& 53.40
& 62.80
\\
Gemini-2.5-pro~\citep{comanici2025gemini}
& 72.50
& 54.60
& 42.40
& 56.50
\\
Claude-Opus-4-1-20250805~\citep{anthropic_claude_2024}
& 75.60
& 56.10
& 58.00
& 63.23
\\

\midrule
\midrule
\ours-Qwen-3-4B (Ours)
& 80.98
& 82.40
& 82.60
& 81.99
\\

\ours-Qwen-2.5-7B (Ours)
& \textbf{93.00}
& \textbf{90.00}
& \textbf{91.00}
& \textbf{91.33}
\\
											
\bottomrule
\end{tabular}
\end{adjustbox}
\caption{Comparison between \ours and baselines in terms of \textbf{Accuracy} across three in-domain datasets. 
The \textsc{OpenRM-Lib-27K} indicates the 27K training data synthesized by the proposed \textsc{OpenRM-Lib} pipeline.}
\label{tab:main}
\vspace{-0.3cm}
\end{table*}

\subsection{Controllable Data Synthesis}\label{sec:data}

A key challenge in training reliable RMs for long-form text evaluation lies in the lack of suitable training data.
Compared to short-form tasks, knowledge-intensive long-form responses are more difficult to collect due to their high information density and the associated annotation costs.
To address this bottleneck, we propose \textsc{OpenRM-Lib}, a straightforward and controllable data synthesis framework that scales the construction of high-quality pairwise training instances for \ours.
Specifically, the core idea is to generate preference pairs by prompting the same LLM to produce responses to the same query under different input conditions.
By controlling the availability of supporting information, such as including or excluding the reference document, we induce a clear and consistent quality gap between the resulting responses. This strategy eliminates the need for manual annotation while ensuring that the resulting preference pairs are both diverse and reliable, making them well-suited for training RMs in knowledge-intensive long-form evaluation settings.

\header{Target-aware Query Generation}
We begin by sampling a set of domain-specific documents (\textit{e.g.,} Wikipedia passages or arXiv papers) and use them as context to prompt a strong LLM  to formulate a \textit{target-aware query} from these documents.
This approach ensures that the generated queries are closely grounded in the source content and can be meaningfully answered with access to the corresponding document.
It is worth noting that a simple question may be associated with multiple relevant documents.
As a result, the synthesized queries are inherently knowledge-intensive and require external evidence for fact checking, providing a solid foundation for constructing high-quality training data for the proposed \ours.

\header{Positive \textit{\&} Negative Pair Synthesis}
Given a query, we generate preference pairs by prompting the LLM under different input conditions.
The \textit{positive response} is generated using both the query and its corresponding reference document, guaranteeing that the answer is grounded, informative, and factually accurate.
In contrast, the \textit{negative response} is generated solely from the query, without access to the reference document, which typically results in incomplete, hallucinated, or less reliable content. 
This contrastive setup introduces a clear quality gap between the two responses, allowing us to construct controllable, scalable pairwise training data tailored to train an agentic RM in \S~\ref{sec:training}.

Finally, we synthesize over 27K high-quality training instances and 2K evaluation instances across three representative scenarios:
(i) Wikipedia, where responses address open-domain questions about specific entities;
(ii) Scientific research, where responses provide technical introductions or surveys; and
(iii) Medical QA, where responses focus on answering health-related questions.


\begin{table}[!t]
\centering
\begin{adjustbox}{width=0.97\textwidth,center}
\setlength\tabcolsep{5.0pt}

\begin{tabular}{@{}p{3cm}  r r | c c  | c c@{}}
\toprule
\textbf{{Methods}} 
& \textbf{Backbone}
& \textbf{Data ($\downarrow$)} 
& \textbf{PandaLM ($\uparrow$)} 
& \textbf{RewardBench ($\uparrow$)} 
& \textbf{Average ($\uparrow$)} 
& \textbf{Average $\Delta$\% ($\uparrow$)} 
\\
\midrule
\rowcolor{Gainsboro} \multicolumn{7}{l}{\bf \em Out-of-domain Evaluation}\\
\midrule

\ours
& Qwen-2.5-7B-Instruct
& \textbf{27K}
& \textbf{79.42}
& 77.66
& \textbf{78.54}
& -
\\

RM-R1
& Qwen-2.5-7B-Instruct
& 72K
& 72.71$_{\downarrow 6.71}$
& 68.34$_{\downarrow 9.32}$
& 70.52
& $\downarrow$11.37\%
\\

JudgeLRM-7B
& Qwen-2.5-7B-Instruct
& 100K 
& 72.37$_{\downarrow 7.05}$
& 74.45$_{\downarrow 3.21}$
& 73.41
& $\downarrow$6.99\%
\\

Prometheus-v2.0
& Mistral-7B-Instruct
& 200K 
& 72.80$_{\downarrow 6.62}$
& 71.55$_{\downarrow 6.11}$
& 72.17
& $\downarrow$8.83\%
\\

RRM-7B
& Qwen-2.5-7B-Instruct
& 420K
& 77.74$_{\downarrow 1.68}$
& \textbf{78.54}$_{\uparrow 0.88}$
& 78.14
& $\downarrow$0.52\%
\\

\bottomrule
\end{tabular}
\end{adjustbox}
\caption{Experiment results in \textbf{Accuracy} on two out-of-domain benchmarks. 
We also report the training data \textit{scale} of each method.
\ours achieves the best average accuracy while using substantially less training data.
}\label{sec:out-of-domain}
\vspace{-0.2cm}

\vspace{-0.1cm}
\end{table}


\begin{figure*}[!t]
    \vskip 0.2in
    \centering
    \includegraphics[width=\linewidth]{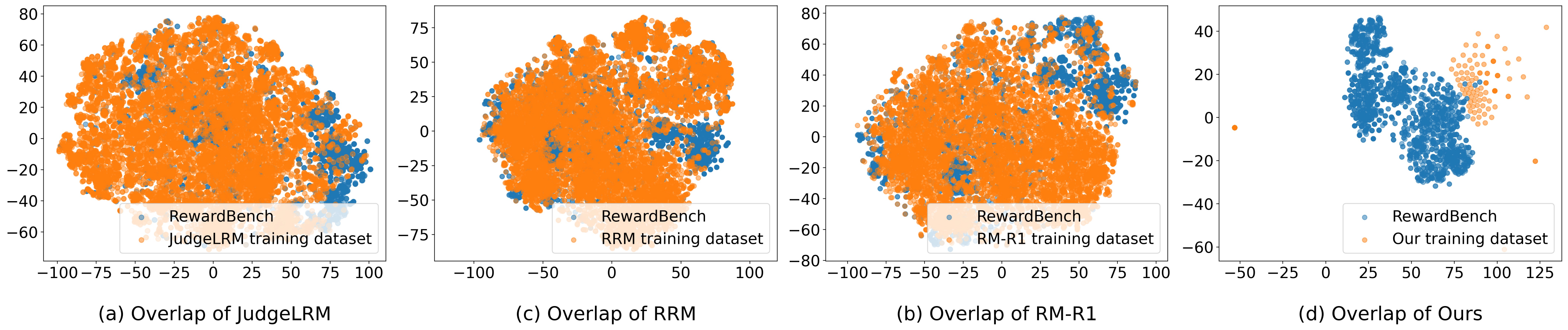}
    \vspace{-0.7cm}
    \caption{Illustration of the overlap between RewardBench and the training datasets of different reward models.
    .}\label{fig:overlap}
    \vspace{-0.4cm}
\end{figure*}

\vspace{-5pt}

\section{Experiment Setup}\label{sec:experiment-setup}
\vspace{-6pt}

\header{Benchmarks}
We evaluate \ours on three newly collected datasets and two widely-used benchmarks.
The three new datasets include (i) 500 Wikipedia QA examples, (ii) 500 scientific research questions, and (iii) 1,000 medical QA examples, respectively.
These datasets are consistent with the training data source, assessing RMs' in-domain performance.
Additionally, we validate \ours on  RewardBench~\cite{lambert2024rewardbench} and PandaLM~\cite{pandalm2024}, validating RMs' out-of-domain performance. 
We report the average accuracy on the test sets.

\header{Baselines}
We consider two types of baselines.

First, \textit{LLM-as-a-Judge}, which prompts or trains a LLM to reason over its internal knowledge and select a better response from multiple candidates.
We include: 
(i) \textit{Skywork-Reward}~\citep{liu2024skywork}, a trained scalar RM;
(ii) \textit{RRM}~\citep{guo2025reward}, a causal judgment framework;
(iii) \textit{JudgeLRM}~\citep{chen2025judgelrm} and \textit{RM-R1}~\citep{chen2025rm}, which are trained via RL to maximize judgment accuracy.
Besides, we also include top-ranked LLMs from the LLM Arena\footnote{\url{https://lmarena.ai/}}, such as Gemini.

Second, \textit{agentic reward modeling}, which augments the LLM-as-Judge approach by integrating external tools, enabling LLMs to use tools on demand.
In more detail, we prompt the LLM to retrieve relevant information using external tools before making a final judgment.
We implement this approach using top-ranked LLMs based on FacTools~\cite{chern2023factool}, a widely used toolkit for tool-augmented fact-checking.
Details of each baseline can be found in Appendix~\ref{sec:app:baselines}.

\header{Implementation Details}
In line with prior work~\cite{chen2025rm, chen2025judgelrm}, we adopt Qwen-2.5-7B-Instruct as the backbone of \ours.
The model can call three tools: Wikipedia Search indexed from the 2018 Wikipedia dump via ColBERT-v2.0~\citep{santhanam2021colbertv2}, arXiv Search provided by LitSearch~\citep{ajith2024litsearch}, and Medical Search over PubMed indexed by MedCPT~\cite{jin2023medcpt}.
Training is conducted for 2 epochs (max prompt length: 4096, batch size: 512) using GRPO with group size 5, clip range 0.5, KL coefficient $\beta=10^{-3}$, and learning rate $10^{-6}$.

\section{Experiment Results}\label{sec:experiment-results}

\subsection{In-domain Evaluation}
Table~\ref{tab:main} compares \ours with baseline RMs on three newly collected test sets. 
We derive the following observations:
(i) LLM-as-a-Judge generally underperform agentic RMs, suggesting that external tools contribute to improved judgment accuracy. However, as most LLMs are not explicitly trained to use tools effectively, prompting them to invoke tools still yields limited performance. 
(ii) Train-based RMs generally underperform commercial LLM-as-judge, likely due to limitations in parameter scale and training methods.

We also continuously train RM-R1 on the same 27k training data as \ours.
However, since RM-R1 is trained with no tool use, its overall performance remains suboptimal, achieving Acc=$68.00\%$. In contrast, by combining \textsc{OpenRM-Lib} with a tailored training method, \ours can autonomously use tools during evaluation and achieve the best accuracy.
It even surpasses large LLM-as-a-judge models such as GPT‑4o and DeepSeek‑R1, demonstrating the effectiveness of our method.
We also provide more discussion in Appendix~\ref{sec:app:discussion}.


\subsection{Out-of-domain Evaluation}

To evaluate generalization ability, we further assess \ours on out-of-domain benchmarks, where the test sets distinct from our training data described in \S~\ref{sec:data}. 
As shown in Table~\ref{sec:out-of-domain}, \ours, despite being trained on only 27K examples, substantially outperforms strong baselines such as RM-R1 and JudgeLRM-7B. 
We attribute this improvement to two main factors. First, the reward model is trained to evaluate knowledge-intensive, long-form responses that demand nuanced reasoning in order to produce accurate judgments. 
This capability contributes to improved generalization across domains.
Second, our agentic modeling framework strengthens the RM's ability to reason over retrieved external evidence.
These two factors equip \ours with notable robustness and the ability to generalize across diverse scenarios.


\begin{figure}[!t]
    \centering
    \includegraphics[width=0.97\linewidth]{figure/reward.pdf}
    \vspace{-0.3cm}
    \caption{Training-time reward curve of the model trained with variant reward functions in \S~\ref{sec:ablation}.
    Since these variants operate on different scales, we normalize them to a unified $[0,1]$ range for fair comparison. 
}\label{fig:reward}
    \vspace{-0.4cm}
    
\end{figure}

\header{A Closer Look at Training Data Overlap}
We find that \ours achieves relatively smaller improvements on RewardBench compared to other benchmarks. To investigate the underlying cause of this discrepancy, we examine the overlap between the training data and the RewardBench test cases.
As shown in Figure~\ref{fig:overlap}, the training sets of several baselines, such as JudgeLRM and RM-R1, have substantial overlap with the RewardBench test cases. 
In contrast, the training data used in our method exhibits only minimal overlap. Nevertheless, \ours still achieves competitive performance despite relying on significantly less and out-of-distribution data. This indicates that the effectiveness of \ours primarily stems from its stronger generalization capability, rather than memorization of seen examples.

\subsection{Ablation Study}\label{sec:ablation}
\vspace{-2pt}

\header{Setup} In \ours, the model is trained with a composite function (Eq.~\ref{eq:reward}), which combines two supervision signals: tool selection $\mathcal{R}_{\text{tool}}$ and judgment accuracy $\mathcal{R}_{\text{EM}}$. 
To better understand the contribution of each, we design three variants for comparison and show the results in Figure~\ref{fig:reward}.

\begin{itemize}
[leftmargin=*,itemsep=2pt,topsep=2pt,parsep=0pt,partopsep=0pt]

    \item \textit{only $\mathcal{R}_{\text{EM}}$}: we use only the judgment accuracy signal \hl{$\mathcal{R}_{\text{EM}} \in [0,1]$}.  
    We observe that removing the tool-selection reward leads to \hl{lazy searching}: the model produces shorter responses with a noticeable performance drop, and avoids tool usage, instead relying on direct prediction, which is inadequate for knowledge-intensive tasks.


    \item \textit{increased} $\mathcal{R}_{\text{tool}}$: the weight of $\mathcal{R}_{\text{tool}}$ is increased from $0.5$ to $1$, yielding \hl{$\mathcal{R}_{\text{EM}} + \text{sign}(\mathcal{R}_{\text{EM}})\cdot \mathcal{R}_{\text{tool}} \in [0,2]$}.  
    We find that model performance is relatively robust to the hyper-parameter $\lambda$. The \textbf{w/ $\mathbf{1}_{\mathcal{R}_{\text{EM}}} \cdot \mathcal{R}_{\text{tool}}$} variant performs comparably to the vanilla model (e.g., $89.60$ vs.\ $86.90$ averaged over three newly collected datasets), with no statistically significant difference under a two-tailed $t$-test. This suggests that the model is not overly sensitive to the weighting of the tool reward.

    \item \textit{decoupled} $\mathcal{R}_{\text{tool}}$: we remove the indicator $\mathbf{1}_{\mathcal{R}_{\text{EM}}}$ when supervising tool usage, formulated as \hl{$\mathcal{R}_{\text{EM}} + \mathcal{R}_{\text{tool}} \in [-0.5,1.5]$}. This decouples tool-use learning from judgment learning, encouraging more explicit tool usage.
    However, Figure~\ref{fig:reward}(d) shows that this decoupling setup leads to reward hacking. 
    \hl{We observe excessive tool usage, \textit{i.e.,} \textit{over-searching}, during both training and evaluation}.
    A likely reason is that the model receives 
    relatively strong positive signals for correct tool usage even when failing to make correct judgments, leading to a form of reward hacking. 
    In contrast, adding an indicator $\mathbf{1}_{\mathcal{R}_{\text{EM}}}$ as a gate in Eq~\ref{eq:reward} can avoid this shortcut.
\end{itemize}

\begin{table*}[!t]

\centering

\begin{adjustbox}{width=0.97\textwidth,center}
\setlength\tabcolsep{4.0pt}

\begin{tabular}{@{}p{5cm} r | c c c  c@{}}
\toprule
\textbf{{Methods}} 
& \textbf{Backbone} 
& \textbf{Self-contain ($\uparrow$)} 
& \textbf{Factuality ($\uparrow$)} 
& \textbf{Average ($\uparrow$)} 
& \textbf{Average $\Delta$\% ($\uparrow$)}

\\

\midrule
\textbf{Metrics} &   &  Accuracy ($\uparrow$)  & Accuracy ($\uparrow$) & Accuracy ($\uparrow$) & Accuracy ($\uparrow$)    \\
\midrule
\ours & Qwen-2.5-7B-Instruct &  2.73  &  2.83 &  2.78 & -  \\
LLM-as-Judge & DeepSeek-V3.1  & 2.67 & 2.47  &  2.57 & $\downarrow$7.55\%   \\
RM-R1~\cite{chen2025rm} & Qwen-2.5-7B-Instruct  & 2.13   & 1.73  &  1.93 &  $\downarrow$30.58\%   \\
\bottomrule
\end{tabular}
\end{adjustbox}
\caption{Human evaluation of different RMs. The \textit{self-contain} metric measures internal consistency of the generated judgments, while \textit{factuality} metric whether the judgments contain counterfactual statements.}
\label{sec:human_eval}
\vspace{-0.25cm}

\end{table*}

\begin{table*}[!t]

\centering
\begin{adjustbox}{width=0.97\textwidth,center}
\setlength\tabcolsep{4.0pt}

\begin{tabular}{@{}p{6cm}  | c c c | c c@{}}
\toprule
\textbf{{Methods}} 
& \textbf{TrutuFulQA ($\uparrow$)} 
& \textbf{MMLU-Pro($\uparrow$)} 
& \textbf{Triviaqa ($\uparrow$)} 
& \textbf{Average ($\uparrow$)} 
& \textbf{Average $\Delta$\% ($\uparrow$)} 
\\
\midrule
\rowcolor{Gainsboro} \multicolumn{6}{l}{\bf \em Out-of-domain Evaluation}\\
\midrule
Qwen-2.5-3B-Instruct & 32.19  &  38.17  &  42.33   &    37.56  &   - \\
$ - w/$ RM-R1  & 32.07  &  38.23  &  43.69   &   37.99  &   $\uparrow$1.14\%\\
$ - w/$ \ours & \textbf{34.03}  &  \textbf{38.60}  &  \textbf{44.56}  &   \textbf{39.03} &   $\uparrow$\textbf{3.91}\%\\
\midrule
Qwen-2.5-7B-Instruct & 39.04  &  48.55  &  50.19   &    45.93  &   -\\
$ - w/$ RM-R1   & 41.00  &  48.81  &  51.53  &   47.11  &   $\uparrow$2.57\%\\
$ - w/$\ours & \textbf{43.33}  &  \textbf{50.63}  &  \textbf{50.12}  &   \textbf{48.03} &   $\uparrow$\textbf{4.57}\%\\


\bottomrule
\end{tabular}
\end{adjustbox}
\caption{Experiment results in terms of accuracy on three out-of-domain benchmarks. We compare Qwen-2.5-3B-Instruct and Qwen-2.5-7B-Instruct trained with either RM-R1 or \ours as reward models under DPO.}
\label{sec:dpo}
\vspace{-0.1cm}

\vspace{-0.2cm}
\end{table*}




\begin{wraptable}{r}{0.48\textwidth}
\centering
\scriptsize
\begin{adjustbox}{width=\linewidth}
\begin{tabular}{@{}p{3.5cm}|c@{}}
\toprule
\textbf{Dataset} & \textbf{UltraFeedback} \\
\midrule
\textbf{Metric} & Accuracy ($\uparrow$) \\
\midrule
\textbf{\ours}  & \textbf{75.18}  \\
RM-R1~\cite{chen2025rm}  & 68.87  \\
\bottomrule
\end{tabular}
\end{adjustbox}
\caption{Selection results accuracy comparing \textbf{\ours} (75.18) and \textbf{RM-R1}~\cite{chen2025rm} (68.87) on the UltraFeedback dataset.}
\label{sec:data_selection}
\vspace{-2mm}
\end{wraptable}


\subsection{Human Evaluation}

To further validate the reliability of \ours, we conduct a human evaluation based on two criteria:
(1) \textbf{Self-containment}, i.e., internal consistency and logical coherence; and
(2) \textbf{Factuality}, i.e., whether the selected response is more factually correct.

Three well-educated volunteers with strong NLP backgrounds evaluate 30 randomly sampled cases from each of the five datasets, using a three-point scale (\textit{3} highest, \textit{1} lowest).
As shown in Table~\ref{sec:human_eval}, \ours outperforms RM-R1 in both criteria, benefiting from incorporating external evidence to produce more reliable judgments.

We also conduct qualitative case studies to analyze its reasoning ability, showing that \ours better orchestrates tool usage for complex tasks. A representative example is provided in Appendix~\ref{sec:appendix}.

\section{Utility Study}\label{sec:application}

We demonstrate the utility of \ours beyond standalone evaluation on the UltraFeedback dataset~\cite{cui2023ultrafeedback}, which contains diverse prompts paired with multiple candidate responses.

\header{\ours for Data Selection}
We use \ours to score and filter training data, reducing supervision noise. As shown in Table~\ref{sec:data_selection}, \ours achieves 75.18\% selection accuracy on UltraFeedback, outperforming RM-R1~\cite{chen2025rm} (68.87\%). Case studies further show that \ours more effectively orchestrates tool usage for complex tasks (Appendix~\ref{sec:app:case} and Table~\ref{tab:case_study}).

\header{\ours for Model Alignment}
We further apply \ours to Direct Preference Optimization (DPO)~\cite{rafailov2023direct} for RLHF~\cite{ouyang2022training}, using the selected positive and negative responses from UltraFeedback as preference pairs. As shown in Table~\ref{sec:dpo}, DPO trained with \ours consistently outperforms the baseline, achieving Acc$=39.03\%$ for the 3B model and Acc$=48.03\%$ for the 7B model on three LLM Arena benchmarks. Compared with RM-R1, \ours also brings larger gains (4.57\% vs. 2.57\%), demonstrating that higher-quality preference data leads to more effective downstream alignment.
\section{Conclusion}
\label{sec:conclusion}

In this paper, we have proposed \ours, a tool-augmented reward model designed to evaluate knowledge-intensive long-form responses. 
\ours aims to reliably assess outputs requiring external grounding and multi-step reasoning. 
By integrating tool usage with reinforcement learning, \ours learns to actively retrieve evidence and make more accurate judgments. 
Extensive experiments across three newly collected datasets and two widely used benchmarks have demonstrated that \ours substantially outperforms existing baselines.
These results highlight the potential of tool-augmented reward modeling to serve as a faithful and scalable evaluator for complex agentic tasks.
Looking ahead, we plan to extend our framework to more complex scenarios involving a broader set of external tools, thereby enhancing its applicability to diverse tasks. 
We also aim to adapt \ours to multimodal settings, enabling models to seamlessly integrate textual reasoning with visual and tabular evidence. 

\newpage
\clearpage



\bibliography{custom}

@article{setlur2024rewarding,
  title={Rewarding progress: Scaling automated process verifiers for llm reasoning},
  author={Setlur, Amrith and Nagpal, Chirag and Fisch, Adam and Geng, Xinyang and Eisenstein, Jacob and Agarwal, Rishabh and Agarwal, Alekh and Berant, Jonathan and Kumar, Aviral},
  journal={arXiv preprint arXiv:2410.08146},
  year={2024}
}

@article{wang2024secrets,
  title={Secrets of rlhf in large language models part ii: Reward modeling},
  author={Wang, Binghai and Zheng, Rui and Chen, Lu and Liu, Yan and Dou, Shihan and Huang, Caishuang and Shen, Wei and Jin, Senjie and Zhou, Enyu and Shi, Chenyu and others},
  journal={arXiv preprint arXiv:2401.06080},
  year={2024}
}

@article{qian2025toolrl,
  title={Toolrl: Reward is all tool learning needs},
  author={Qian, Cheng and Acikgoz, Emre Can and He, Qi and Wang, Hongru and Chen, Xiusi and Hakkani-T{\"u}r, Dilek and Tur, Gokhan and Ji, Heng},
  journal={arXiv preprint arXiv:2504.13958},
  year={2025}
}

@article{xie2025capo,
  title={CAPO: Towards Enhancing LLM Reasoning through Generative Credit Assignment},
  author={Xie, Guofu and Shi, Yunsheng and Tian, Hongtao and Yao, Ting and Zhang, Xiao},
  journal={arXiv preprint arXiv:2508.02298},
  year={2025}
}

@article{bercovich2025llama,
  title={Llama-nemotron: Efficient reasoning models},
  author={Bercovich, Akhiad and Levy, Itay and Golan, Izik and Dabbah, Mohammad and El-Yaniv, Ran and Puny, Omri and Galil, Ido and Moshe, Zach and Ronen, Tomer and Nabwani, Najeeb and others},
  journal={arXiv preprint arXiv:2505.00949},
  year={2025}
}

@article{zhang2025rlvmr,
  title={Rlvmr: Reinforcement learning with verifiable meta-reasoning rewards for robust long-horizon agents},
  author={Zhang, Zijing and Chen, Ziyang and Li, Mingxiao and Tu, Zhaopeng and Li, Xiaolong},
  journal={arXiv preprint arXiv:2507.22844},
  year={2025}
}

@inproceedings{li2025generation,
  title={From generation to judgment: Opportunities and challenges of llm-as-a-judge},
  author={Li, Dawei and Jiang, Bohan and Huang, Liangjie and Beigi, Alimohammad and Zhao, Chengshuai and Tan, Zhen and Bhattacharjee, Amrita and Jiang, Yuxuan and Chen, Canyu and Wu, Tianhao and others},
  booktitle={Proceedings of the 2025 Conference on Empirical Methods in Natural Language Processing},
  pages={2757--2791},
  year={2025}
}

@inproceedings{kim2023prometheus,
  title={Prometheus: Inducing fine-grained evaluation capability in language models},
  author={Kim, Seungone and Shin, Jamin and Cho, Yejin and Jang, Joel and Longpre, Shayne and Lee, Hwaran and Yun, Sangdoo and Shin, Seongjin and Kim, Sungdong and Thorne, James and others},
  booktitle={The Twelfth International Conference on Learning Representations},
  year={2023}
}

@article{li2023generative,
  title={Generative judge for evaluating alignment},
  author={Li, Junlong and Sun, Shichao and Yuan, Weizhe and Fan, Run-Ze and Zhao, Hai and Liu, Pengfei},
  journal={arXiv preprint arXiv:2310.05470},
  year={2023}
}

@article{shao2024deepseekmath,
  title={Deepseekmath: Pushing the limits of mathematical reasoning in open language models},
  author={Shao, Zhihong and Wang, Peiyi and Zhu, Qihao and Xu, Runxin and Song, Junxiao and Bi, Xiao and Zhang, Haowei and Zhang, Mingchuan and Li, YK and Wu, Yang and others},
  journal={arXiv preprint arXiv:2402.03300},
  year={2024}
}

@article{wen2025reinforcement,
  title={Reinforcement learning with verifiable rewards implicitly incentivizes correct reasoning in base llms},
  author={Wen, Xumeng and Liu, Zihan and Zheng, Shun and Ye, Shengyu and Wu, Zhirong and Wang, Yang and Xu, Zhijian and Liang, Xiao and Li, Junjie and Miao, Ziming and others},
  journal={arXiv preprint arXiv:2506.14245},
  year={2025}
}

@article{zeng2025simplerl,
  title={Simplerl-zoo: Investigating and taming zero reinforcement learning for open base models in the wild},
  author={Zeng, Weihao and Huang, Yuzhen and Liu, Qian and Liu, Wei and He, Keqing and Ma, Zejun and He, Junxian},
  journal={arXiv preprint arXiv:2503.18892},
  year={2025}
}

@article{ouyang2022training,
  title={Training language models to follow instructions with human feedback},
  author={Ouyang, Long and Wu, Jeffrey and Jiang, Xu and Almeida, Diogo and Wainwright, Carroll and Mishkin, Pamela and Zhang, Chong and Agarwal, Sandhini and Slama, Katarina and Ray, Alex and others},
  journal={Advances in neural information processing systems},
  volume={35},
  pages={27730--27744},
  year={2022}
}

@article{liu2025trust,
  title={Trust, But Verify: A Self-Verification Approach to Reinforcement Learning with Verifiable Rewards},
  author={Liu, Xiaoyuan and Liang, Tian and He, Zhiwei and Xu, Jiahao and Wang, Wenxuan and He, Pinjia and Tu, Zhaopeng and Mi, Haitao and Yu, Dong},
  journal={arXiv preprint arXiv:2505.13445},
  year={2025}
}

@inproceedings{song2024preference,
  title={Preference ranking optimization for human alignment},
  author={Song, Feifan and Yu, Bowen and Li, Minghao and Yu, Haiyang and Huang, Fei and Li, Yongbin and Wang, Houfeng},
  booktitle={Proceedings of the AAAI Conference on Artificial Intelligence},
  volume={38},
  pages={18990--18998},
  year={2024}
}

@article{chen2025rm,
  title={Rm-r1: Reward modeling as reasoning},
  author={Chen, Xiusi and Li, Gaotang and Wang, Ziqi and Jin, Bowen and Qian, Cheng and Wang, Yu and Wang, Hongru and Zhang, Yu and Zhang, Denghui and Zhang, Tong and others},
  journal={arXiv preprint arXiv:2505.02387},
  year={2025}
}

@article{guo2025reward,
  title={Reward reasoning model},
  author={Guo, Jiaxin and Chi, Zewen and Dong, Li and Dong, Qingxiu and Wu, Xun and Huang, Shaohan and Wei, Furu},
  journal={arXiv preprint arXiv:2505.14674},
  year={2025}
}

@article{zhu2025deepreview,
  title={Deepreview: Improving llm-based paper review with human-like deep thinking process},
  author={Zhu, Minjun and Weng, Yixuan and Yang, Linyi and Zhang, Yue},
  journal={arXiv preprint arXiv:2503.08569},
  year={2025}
}

@article{liu2024skywork,
  title={Skywork-reward: Bag of tricks for reward modeling in llms},
  author={Liu, Chris Yuhao and Zeng, Liang and Liu, Jiacai and Yan, Rui and He, Jujie and Wang, Chaojie and Yan, Shuicheng and Liu, Yang and Zhou, Yahui},
  journal={arXiv preprint arXiv:2410.18451},
  year={2024}
}

@article{gu2024survey,
  title={A survey on llm-as-a-judge},
  author={Gu, Jiawei and Jiang, Xuhui and Shi, Zhichao and Tan, Hexiang and Zhai, Xuehao and Xu, Chengjin and Li, Wei and Shen, Yinghan and Ma, Shengjie and Liu, Honghao and others},
  journal={arXiv preprint arXiv:2411.15594},
  year={2024}
}

@article{zheng2023judging,
  title={Judging llm-as-a-judge with mt-bench and chatbot arena},
  author={Zheng, Lianmin and Chiang, Wei-Lin and Sheng, Ying and Zhuang, Siyuan and Wu, Zhanghao and Zhuang, Yonghao and Lin, Zi and Li, Zhuohan and Li, Dacheng and Xing, Eric and others},
  journal={Advances in neural information processing systems},
  volume={36},
  pages={46595--46623},
  year={2023}
}

@article{li2024llms,
  title={Llms-as-judges: a comprehensive survey on llm-based evaluation methods},
  author={Li, Haitao and Dong, Qian and Chen, Junjie and Su, Huixue and Zhou, Yujia and Ai, Qingyao and Ye, Ziyi and Liu, Yiqun},
  journal={arXiv preprint arXiv:2412.05579},
  year={2024}
}

@misc{wang2024helpsteer2,
      title={HelpSteer2: Open-source dataset for training top-performing reward models}, 
      author={Zhilin Wang and Yi Dong and Olivier Delalleau and Jiaqi Zeng and Gerald Shen and Daniel Egert and Jimmy J. Zhang and Makesh Narsimhan Sreedhar and Oleksii Kuchaiev},
      year={2024},
      eprint={2406.08673},
      archivePrefix={arXiv},
}

@article{chen2024travelagent,
  title={Travelagent: An ai assistant for personalized travel planning},
  author={Chen, Aili and Ge, Xuyang and Fu, Ziquan and Xiao, Yanghua and Chen, Jiangjie},
  journal={arXiv preprint arXiv:2409.08069},
  year={2024}
}

@inproceedings{wang2024executable,
  title={Executable code actions elicit better llm agents},
  author={Wang, Xingyao and Chen, Yangyi and Yuan, Lifan and Zhang, Yizhe and Li, Yunzhu and Peng, Hao and Ji, Heng},
  booktitle={Forty-first International Conference on Machine Learning},
  year={2024}
}

@article{wei2024long,
  title={Long-form factuality in large language models},
  author={Wei, Jerry and Yang, Chengrun and Song, Xinying and Lu, Yifeng and Hu, Nathan and Huang, Jie and Tran, Dustin and Peng, Daiyi and Liu, Ruibo and Huang, Da and others},
  journal={Advances in Neural Information Processing Systems},
  volume={37},
  pages={80756--80827},
  year={2024}
}

@article{jin2025search,
  title={Search-r1: Training llms to reason and leverage search engines with reinforcement learning},
  author={Jin, Bowen and Zeng, Hansi and Yue, Zhenrui and Yoon, Jinsung and Arik, Sercan and Wang, Dong and Zamani, Hamed and Han, Jiawei},
  journal={arXiv preprint arXiv:2503.09516},
  year={2025}
}

@article{liu2024deepseek,
  title={Deepseek-v3 technical report},
  author={Liu, Aixin and Feng, Bei and Xue, Bing and Wang, Bingxuan and Wu, Bochao and Lu, Chengda and Zhao, Chenggang and Deng, Chengqi and Zhang, Chenyu and Ruan, Chong and others},
  journal={arXiv preprint arXiv:2412.19437},
  year={2024}
}

@article{hurst2024gpt,
  title={Gpt-4o system card},
  author={Hurst, Aaron and Lerer, Adam and Goucher, Adam P and Perelman, Adam and Ramesh, Aditya and Clark, Aidan and Ostrow, AJ and Welihinda, Akila and Hayes, Alan and Radford, Alec and others},
  journal={arXiv preprint arXiv:2410.21276},
  year={2024}
}

@article{comanici2025gemini,
  title={Gemini 2.5: Pushing the frontier with advanced reasoning, multimodality, long context, and next generation agentic capabilities},
  author={Comanici, Gheorghe and Bieber, Eric and Schaekermann, Mike and Pasupat, Ice and Sachdeva, Noveen and Dhillon, Inderjit and Blistein, Marcel and Ram, Ori and Zhang, Dan and Rosen, Evan and others},
  journal={arXiv preprint arXiv:2507.06261},
  year={2025}
}

@misc{anthropic_claude_2024,
  author       = {{Anthropic Team}},
  title        = {{Claude Opus 4.1}},
  howpublished = {Anthropic News, \url{https://www.anthropic.com/news/claude-opus-4-1}},
  month        = {August},
  year         = {2025},
  note         = {Accessed: August 06, 2025},
}

@article{chen2025judgelrm,
  title={Judgelrm: Large reasoning models as a judge},
  author={Chen, Nuo and Hu, Zhiyuan and Zou, Qingyun and Wu, Jiaying and Wang, Qian and Hooi, Bryan and He, Bingsheng},
  journal={arXiv preprint arXiv:2504.00050},
  year={2025}
}

@misc{whitehouse2025j1incentivizingthinkingllmasajudge,
      title={J1: Incentivizing Thinking in LLM-as-a-Judge via Reinforcement Learning}, 
      author={Chenxi Whitehouse and Tianlu Wang and Ping Yu and Xian Li and Jason Weston and Ilia Kulikov and Swarnadeep Saha},
      year={2025},
      eprint={2505.10320},
      archivePrefix={arXiv},
      primaryClass={cs.CL},
      url={https://arxiv.org/abs/2505.10320}, 
}

@misc{li2024llmsasjudgescomprehensivesurveyllmbased,
      title={LLMs-as-Judges: A Comprehensive Survey on LLM-based Evaluation Methods}, 
      author={Haitao Li and Qian Dong and Junjie Chen and Huixue Su and Yujia Zhou and Qingyao Ai and Ziyi Ye and Yiqun Liu},
      year={2024},
      eprint={2412.05579},
      archivePrefix={arXiv},
      primaryClass={cs.CL},
      url={https://arxiv.org/abs/2412.05579}, 
}

@article{lambert2024rewardbench,
  title={Rewardbench: Evaluating reward models for language modeling},
  author={Lambert, Nathan and Pyatkin, Valentina and Morrison, Jacob and Miranda, LJ and Lin, Bill Yuchen and Chandu, Khyathi and Dziri, Nouha and Kumar, Sachin and Zick, Tom and Choi, Yejin and others},
  journal={arXiv preprint arXiv:2403.13787},
  year={2024}
}

@article{pandalm2024,
      title={PandaLM: An Automatic Evaluation Benchmark for LLM Instruction Tuning Optimization}, 
      author={Wang, Yidong and Yu, Zhuohao and Zeng, Zhengran and Yang, Linyi and Wang, Cunxiang and Chen, Hao and Jiang, Chaoya and Xie, Rui and Wang, Jindong and Xie, Xing and Ye, Wei and Zhang, Shikun and Zhang, Yue},
      booktitle={International Conference on Learning Representations (ICLR)},
      year={2024}
}

@article{ajith2024litsearch,
  title={Litsearch: A retrieval benchmark for scientific literature search},
  author={Ajith, Anirudh and Xia, Mengzhou and Chevalier, Alexis and Goyal, Tanya and Chen, Danqi and Gao, Tianyu},
  journal={arXiv preprint arXiv:2407.18940},
  year={2024}
}

@article{santhanam2021colbertv2,
  title={Colbertv2: Effective and efficient retrieval via lightweight late interaction},
  author={Santhanam, Keshav and Khattab, Omar and Saad-Falcon, Jon and Potts, Christopher and Zaharia, Matei},
  journal={arXiv preprint arXiv:2112.01488},
  year={2021}
}

@article{rafailov2023direct,
  title={Direct preference optimization: Your language model is secretly a reward model},
  author={Rafailov, Rafael and Sharma, Archit and Mitchell, Eric and Manning, Christopher D and Ermon, Stefano and Finn, Chelsea},
  journal={Advances in neural information processing systems},
  volume={36},
  pages={53728--53741},
  year={2023}
}

@article{chern2023factool,
  title={FacTool: Factuality Detection in Generative AI--A Tool Augmented Framework for Multi-Task and Multi-Domain Scenarios},
  author={Chern, I and Chern, Steffi and Chen, Shiqi and Yuan, Weizhe and Feng, Kehua and Zhou, Chunting and He, Junxian and Neubig, Graham and Liu, Pengfei and others},
  journal={arXiv preprint arXiv:2307.13528},
  year={2023}
}

@article{cui2023ultrafeedback,
  title={Ultrafeedback: Boosting language models with scaled ai feedback},
  author={Cui, Ganqu and Yuan, Lifan and Ding, Ning and Yao, Guanming and He, Bingxiang and Zhu, Wei and Ni, Yuan and Xie, Guotong and Xie, Ruobing and Lin, Yankai and others},
  journal={arXiv preprint arXiv:2310.01377},
  year={2023}
}

@article{li2023tool,
  title={Tool-augmented reward modeling},
  author={Li, Lei and Chai, Yekun and Wang, Shuohuan and Sun, Yu and Tian, Hao and Zhang, Ningyu and Wu, Hua},
  journal={arXiv preprint arXiv:2310.01045},
  year={2023}
}

@article{zheng2025deepresearcher,
  title={Deepresearcher: Scaling deep research via reinforcement learning in real-world environments},
  author={Zheng, Yuxiang and Fu, Dayuan and Hu, Xiangkun and Cai, Xiaojie and Ye, Lyumanshan and Lu, Pengrui and Liu, Pengfei},
  journal={arXiv preprint arXiv:2504.03160},
  year={2025}
}

@article{jin2023medcpt,
  title={Medcpt: Contrastive pre-trained transformers with large-scale pubmed search logs for zero-shot biomedical information retrieval},
  author={Jin, Qiao and Kim, Won and Chen, Qingyu and Comeau, Donald C and Yeganova, Lana and Wilbur, W John and Lu, Zhiyong},
  journal={Bioinformatics},
  volume={39},
  number={11},
  pages={btad651},
  year={2023},
  publisher={Oxford University Press}
}
\bibliographystyle{colm2026_conference}

\appendix

\onecolumn

\section{Appendix}\label{sec:appendix}


\begin{table*}[!t]

\centering
\setlength{\tabcolsep}{2.5pt}
\begin{adjustbox}{width=\columnwidth,center}
\begin{tabular}{@{}p{6cm} c c c c c}
\toprule
\textbf{Statistic}   
& \begin{tabular}{c}
        \textbf{\ours}  \\ (Ours)
    \end{tabular}
&  \begin{tabular}{c}
        \textbf{SkyWork-Reward}  \\ \cite{liu2024skywork}
    \end{tabular}
&  \begin{tabular}{c}
        \textbf{TARA}  \\ \cite{li2023tool}
    \end{tabular}
& \begin{tabular}{c}
        \textbf{PandaLM}  \\ \cite{pandalm2024}
    \end{tabular}
& \begin{tabular}{c}
        \textbf{RM-R1}  \\ \cite{chen2025rm}
    \end{tabular}
\\

\hline
\rowcolor{green!12} \multicolumn{6}{l}{\# Reward Modeling Perspective}\\

$ $ - \textit{Evidence-grounded Judgment}
& \ding{52}
& \ding{56}
& \ding{52}
& \ding{56}
& \ding{56}
\\

$ $ - \textit{Reasoning for Step-by-Step Judgment}
& \ding{52}
& \ding{56}
& \ding{56}
& \ding{56}
& \ding{52}
\\

$ $ - \textit{Language-based Explanation}
& \ding{52}
& \ding{56}
& \ding{52}
& \ding{52}
& \ding{52}
\\

\hline
\rowcolor{blue!12} \multicolumn{6}{l}{\# How to Train the RMs}\\

$ $ - \textit{Question Construction}
& Synthesized
& Handcrafted
& Synthesized
& Synthesized
& Synthesized
\\

$ $ - \textit{Candidate Responses Construction}
& Synthesized
& Handcrafted
& Synthesized
& Synthesized
& Synthesized
\\

$ $ - \textit{Binary Label Annotation}
& Unsupervised
& LLM-as-a-Judge
& LLM-as-a-Judge
& LLM-as-a-Judge
& LLM-as-a-Judge
\\

$ $ - \textit{Training Paradigm}
& RLVR
& Contrastive Learning
& SFT
& SFT
& RLVR
\\

\bottomrule
\end{tabular}
\end{adjustbox}
\vspace{-0.1cm}
\caption{Comparisons between our method and the previous reward modeling approach. The SFT indicates supervised fine-tuning, while the RLVR indicates reinforcement learning with verifiable reward.
The binary indicates distinguishing the positive (i.e., 1) and negative (i.e., 0) responses from the candidate responses.}\label{tab:comparison}
\end{table*}

\subsection{Discussion}\label{sec:app:discussion}

Below, we provide further discussion on the design choices and implications of \ours.

\subsubsection{Question: Why augmenting RMs with search tools matters?}
A fundamental limitation of existing RMs lies in their reliance on the model’s internal parametric knowledge when evaluating long-form, knowledge-intensive outputs. In many real-world agentic tasks—such as scientific writing, medical question answering, and deep research reports—judging response quality often requires verifying factual claims against external and up-to-date information sources, which cannot be reliably encoded in model parameters alone. 

Augmenting RMs with search tools enables them to actively retrieve relevant evidence during evaluation, thereby grounding their judgments in external knowledge rather than surface-level cues such as fluency, structure, or verbosity. This capability is particularly critical for long-form evaluation, where errors may be subtle, distributed across multiple claims, and difficult to detect without explicit fact checking. By empowering RMs to access external information on demand, \ours bridges the gap between preference modeling and evidence-based verification, leading to more reliable and interpretable reward signals for complex agentic tasks.

\subsubsection{Question: How to Extend the Scope of Tools in \textsc{OpenReward}.}
In this work, we focus on search tools with a simple and widely adopted interface, where the model only needs to generate a query to retrieve relevant documents. This design choice is motivated by two considerations. First, information retrieval constitutes the most common and fundamental external capability required for evaluating knowledge-intensive text. Second, a lightweight search interface reduces engineering complexity and lowers the barrier for stable reinforcement learning, allowing the RM to focus on learning when and how to retrieve evidence rather than struggling with tool-specific execution details.

Nevertheless, the proposed framework is not inherently limited to search-only tools. In future work, we plan to extend \ours to broader tool ecosystems, including code interpreters, Web APIs (e.g., RESTful services), and GUI-based tools. Such extensions would allow RMs to verify numerical computations, interact with structured databases, or validate claims through executable procedures, further enhancing their applicability to more diverse and complex evaluation scenarios.

\subsubsection{Potential Reward Hacking in Training \textsc{OpenReward}.}

Reward hacking is a well-known challenge in reinforcement learning, particularly when intermediate rewards are introduced to guide multi-step behaviors. In \ours, we explicitly address this risk through a gated reward design, where tool-use rewards are only activated when the final judgment is correct. 
This gated mechanism discourages degenerate strategies such as excessive or irrelevant tool calls that do not contribute to accurate evaluation. Empirically, our ablation studies show that removing this gating leads to over-searching behaviors, while relying solely on final outcome rewards results in lazy searching, confirming the necessity of carefully coupling intermediate and terminal supervision.

While our results indicate that \ours effectively mitigates reward hacking under the current training setup, we do not claim that the problem is fully eliminated. Understanding the precise mechanisms by which RMs balance evidence acquisition and decision making remains an open research question. Future work may investigate more principled credit assignment strategies, adaptive tool budgets, or theoretical analyses of reward shaping to further improve robustness in training agentic RMs.

\subsection{Runtime Case of \ours}\label{sec:app:case}

We provide a concrete runtime case of example in Table~\ref{tab:case_study}.
The output example of a strong baseline, i.e., RM-R1, are also provided for an intuitive comparison.

\begin{table*}[!t]

\label{tab:case_study}
\centering
\begin{adjustbox}{width=\columnwidth,center}
\begin{tabular}{@{}p{\columnwidth}@{}}
\toprule
\textbf{Dataset:} UltraFeedback \\
\midrule
\textbf{Question:} How have foreign influences played a role in shaping Romania's cultural identity? \\
\midrule
\textbf{Answer A:} Foreign influences have played an important role in shaping Romania's cultural identity. Due to its geographic location, Romania has been exposed to various cultures for centuries, from the neighboring Hungarians and Austrians to further away Ottomans and Russians. As a result, Romanian culture has incorporated aspects from all of these cultures... For example, the traditional Romanian New Year's Eve dinner includes ntbura... \\
\midrule
\textbf{Answer B:} The influence of foreign cultures has played a significant role in shaping Romania's cultural identity throughout history. Here are some examples: 1. Greek Influence: Ancient Greece had a profound impact on Romanian culture... 2. Ottoman Influence... 3. German Influence... 4. French Influence... 5. Italian Influence... \\
\midrule
\textbf{RM-R1 Judgment:} \\
\begin{minipage}[t]{\columnwidth}
Answer B is better due to its systematic, \fcolorbox{red}{white}{numbered breakdown} of five specific cultural influences (Greek, Ottoman, etc.). This provides a more complete and \fcolorbox{red}{white}{structured} inventory than Answer A's general narrative, enhancing clarity. This organization is more useful for a user seeking a \fcolorbox{red}{white}{comprehensive list} and easy-to-digest information.
$<answer>$B$</answer>$ \\[4pt]
\textbf{\textcolor{red}{\ding{55}\ Wrong answer: the explanation for each response is brief and lacks in-depth analysis, which only focus on the structure while neglecting content factuality.}}
\end{minipage} \\[10pt]
\midrule
\textbf{\ours Judgment:} \\
\begin{minipage}[t]{\columnwidth}
Both answers provide information about the role of foreign influences in shaping Romania's cultural identity. However, Answer A is \fcolorbox{green}{white}{more detailed, providing concrete examples} such as... I will perform a search to see if there are any critical steps missed by either answer, before making a final decision. $<$search$>$WIKI(foreign influences on Romanian culture)$<$\/search$>$... \fcolorbox{green}{white}{The search results} returned content related to Slavic influences on Romanian culture, \fcolorbox{green}{white}{consistent with the description in Answer A.} Answer A is more detailed and specific, providing concrete examples such as ... Answer B, while informative, is more structured as a list of examples and \fcolorbox{green}{white}{lacks the same level of detail and specificity}. Therefore, Answer A is more comprehensive and better answers the input question.
$<$answer$>$A$<$\/answer$>$ \\[4pt]
\textbf{\textcolor{green!60!black}{\ding{51}\ Correct answer: a search tool is invoked to assist judgment. The model eventually identifies that the content of Answer A is more detailed than Answer B.}}
\end{minipage} \\

\bottomrule
\end{tabular}
\end{adjustbox}
\caption{A showcase comparing the judgments of RM-R1 and \ours on a representative case.}
\end{table*}

\subsection{LLM Usage}

In this work, large language models (LLMs) were not used for research ideation, methodological development, or the generation of any original scientific content.
LLMs were employed solely in a limited, auxiliary capacity as general-purpose tools for grammar checking and minor wording refinement during manuscript preparation.
All conceptual formulation, experimental design, implementation, empirical analysis, and the writing of the manuscript’s content were conducted entirely by the authors.

\subsection{Baseline Implementation Details}
\label{sec:app:baselines}

We provide additional details on the baseline RMs evaluated in our experiments.

\paragraph{Train-based RMs.}
These baselines are directly trained to predict preference scores or rank candidate responses.  
\textbf{Skywork-Reward}~\citep{liu2024skywork} and \textbf{JudgeLRM}~\citep{chen2025judgelrm} are trained using large-scale human preference datasets such as UltraFeedback and JudgeLM.  
\textbf{RRM}~\citep{guo2025reward} adopts a multi-stage preference optimization pipeline with iterative RM updates.  
\textbf{RM-R1}~\citep{chen2025rm} is trained with reinforcement learning signals derived from rule-based reward functions.  
For all train-based RMs, we use the 7B versions released and follow their official evaluation setups.  

\paragraph{LLM-as-Judge.}
In this category, large language models are prompted to act as judges without external retrieval or verification tools.  
We select top-ranked models from the \textit{Arena Leaderboard}, including \textbf{DeepSeek-V3.1}, \textbf{GPT-4o}, \textbf{Gemini-2.5-Pro} and \textbf{ Claude-Opus-4}.  
Each model receives the same pairwise comparison prompt as \ours, and outputs are parsed into binary preferences(A or B).  
Temperature is set to 0.3 for deterministic judgment generation.

\paragraph{Agentic RMs.}
These methods extend the LLM-as-Judge paradigm by incorporating external tools.  
We adopt \textbf{FacTool}~\citep{chern2023factool}, a popular toolkit for tool-augmented fact-checking and evidence retrieval.  
Specifically, the model first retrieves relevant documents using external tools like WikiSearch and ArxivSearch, then re-evaluates candidate responses conditioned on the retrieved evidence.  
We test this approach with top-ranked models mentioned in LLM-as-Judge.

\paragraph{Evaluation Protocol.}
All baselines are evaluated on the same preference alignment datasets as \ours, using accuracy as the main metric (i.e. whether the predicted preference matches the ground-truth annotation).  
Each judgment is computed independently without majority voting, and the models are evaluated under identical sampling configurations.

\subsection{Prompt of \ours}\label{sec:app:details}

\begin{EnvBox}[System Prompt Template of \ours]
\begin{lstlisting}
You are an impartial judge tasked with evaluating two candidate responses and determining which one better answers the input question with higher quality. Your evaluation should focus on factual correctness, clarity, completeness, and helpfulness.

In judging the response, you must start **concise** reasoning inside <think> and </think>. Whenever you want to ground your judgement with external evidence, you are ENCOURAGED to pause the reasoning, and then activtely **search for reference** from relevant corpora inside "<search>" and "</search>". The search results will be enclosed and returned in <information> and </information>. After obtaining the results, you can continue your reasoning.

Note: For research-oriented questions, search for relevant scientific papers on arXiv with <search>ARXIV("YOUR QUERY")</search>; for everyday open-domain questions, search Wikipedia with <search>WIKI("YOUR QUERY")</search>. Since retrieval is commonly imperfect, you should always try more DIFFERENT ISSUES when failing to find useful information. But if you still fail after 2 attempts, you should terminate the search process and continue to reason the final judgment. 

Starting below, you should interleave "<think> </think>" and "<search> </search>" multiple times to progressively gather relevant evidence in a step-by-step manner. After gathering enough evidence, you must conclude your final judgment inside <think> and </think>, followed by your final choice (ONLY 'A' or 'B') inside <answer> and </answer>.

\end{lstlisting}
\end{EnvBox}

\begin{EnvBox}[User Prompt of \ours]
\begin{lstlisting}
Please help me Judge which of the two candidate responses better answers the question: {question}

Answer A: {answer a}

Answer B: {answer b}

NOTE: Using relevant evidence from the search results to support your judgment, focusing on **factual correctness**, **clarity**, **completeness**, and **helpfulness**. Trivial factors such as the order of the responses, their length, or writing style should not influence your decision. Do not infer polarity from position: neither the first response nor the second is inherently positive or negative.
\end{lstlisting}
\end{EnvBox}

\begin{EnvBox}[Input Question and Two Candidate Prompts]
\begin{lstlisting}
Question: How does the use of the Balanced Winnow classifier, in conjunction with a novel feature set combining lemmatized words, stylometric features, and world knowledge features, address the specific challenges of a fine-grained, multi-label genre classification task for French literary texts, as opposed to more traditional classifiers and broader genre definitions used in foundational studies like those by Kessler et al. (1997)?

Answer A: The provided information does not discuss the Balanced Winnow classifier, genre classification, or French literary texts. Therefore, I will address the question based on my knowledge.
The use of the Balanced Winnow classifier, in conjunction with a novel feature set combining lemmatized words, stylometric features, and world knowledge features, addresses the specific challenges of fine-grained, multi-label genre classification for French literary texts by leveraging its ability to handle high-dimensional feature spaces and its effectiveness in binary and multi-label classification tasks. Unlike traditional classifiers such as Naive Bayes or decision trees, which may struggle with the complexity and nuance required for fine-grained distinctions, balanced Winnow - a multiplicative weight-update algorithm-excels in environments where feature relevance is sparse but critical, making it well-suited for literary analysis where discriminative features might be subtle and varied. The novel feature set enhances this capability: lemmatized words help normalize morphological variations, capturing core lexical patterns without noise from inflectional changes; stylometric features (e.g., sentence length, lexical richness, syntactic structures) quantify authorial and stylistic nuances that are pivotal in distinguishing literary genres; and world knowledge features (e.g., historical context, thematic elements derived from external databases) incorporate semantic and cultural depth, addressing limitations of purely statistical approaches. This combination allows the model to capture both surface-level patterns and deeper semantic cues, which is essential for multi-label tasks where texts often exhibit hybrid genre characteristics.
In contrast, foundational studies like Kessler et al. (1997) often relied on broader genre definitions (e.g., fiction vs. non-fiction) and simpler feature sets (e.g., raw word frequencies) with classifiers like Naive Bayes, which prioritize probabilistic coherence over nuanced feature interactions. These approaches are less adaptable to fine-grained categories (e.g., subgenres like "symbolist poetry" or "existentialist novel") and multi-label scenarios, as they may oversimplify linguistic and extralinguistic cues. The Balanced Winnow classifier, with its iterative weight adjustment and emphasis on informative features, reduces error propagation in complex label sets and improves precision in identifying overlapping genres, thus offering a more robust framework for the intricacies of French literary texts. 

Answer B: The DEFT 2014 challenge presented a particularly nuanced genre classification task: fine-grained categorization of French literary texts, where documents could belong to multiple genres simultaneously. This differs significantly from the broader genre definitions used in foundational studies, which often distinguished between fundamentally different text types (e.g., prose vs. poetry, written vs. spoken language). The task here required distinguishing between subtler distinctions within the same literary domain, making it both multi-label and fine-grained. The choice of the Balanced Winnow classifier, combined with a novel feature set, directly addresses these specific challenges. While traditional classifiers might struggle with the high dimensionality and feature sparsity common in text classification, Balanced Winnow is well-suited for such scenarios. It is an online linear classifier related to the Perceptron, but designed to handle a large number of features efficiently-a critical capability when combining diverse feature types. The novel feature set itself is key to capturing the fine-grained nuances of literary genres. Lemmatized words help standardize the vocabulary to capture core thematic content (e.g., recognizing that "crime" and "crimes" convey the same theme relevant to a police novel). Stylometric features are essential for detecting structural differences, such as the varied use of punctuation or sentence length that might distinguish poetry from prose or even one type of prose from another. World knowledge features incorporate a deeper understanding beyond surface-level text, potentially capturing context or connotations that are genre-specific.
By combining these features, the approach captures both content (lemmatized words, world knowledge) and style (stylometric features), which is necessary because literary genres often depend on a blend of these elements. For instance, two genres might share similar themes but differ significantly in stylistic presentation. The multi-label aspect is addressed through the classifier's ability to manage multiple, potentially overlapping categories by not forcing mutually exclusive decisions.
In contrast, the broader genre definitions in earlier work might have relied on more obvious distinguishing features, making simpler classifiers or single feature types sufficient. The fine-grained, multi-label nature of the DEFT task demands a more sophisticated combination of feature engineering and a robust classifier like Balanced Winnow to achieve high precision without sacrificing computational efficiency.
\end{lstlisting}
\end{EnvBox}

\end{document}